\documentclass{article}

\PassOptionsToPackage{table}{xcolor}
\usepackage{iclr2027_conference,times}
\usepackage[utf8]{inputenc} % allow utf-8 input
\usepackage[T1]{fontenc}    % use 8-bit T1 fonts
\usepackage{hyperref}       % hyperlinks
\usepackage{url}            % simple URL typesetting
\usepackage[table]{xcolor}  % colors and table shading
\usepackage{caption}
\usepackage{booktabs}       % professional-quality tables
\usepackage{amsfonts}       % blackboard math symbols
\usepackage{nicefrac}       % compact symbols for 1/2, etc.
\usepackage{microtype}      % microtypography
\usepackage{graphicx}
\usepackage{amsmath}
\usepackage{amssymb}
\usepackage{mathtools}
\usepackage{amsthm}
\usepackage[capitalize,noabbrev]{cleveref}
\usepackage{latexsym}
\usepackage{multirow}
\usepackage{longtable}
\usepackage[english]{babel}
\usepackage{bm}
\usepackage{enumitem}
\usepackage{wrapfig}
\usepackage{xspace}
\usepackage{fontawesome5}
\usepackage{needspace}
\usepackage{etoolbox}
\NewCommandCopy{\AnyActOriginalSubsection}{\subsection}
\RenewDocumentCommand{\subsection}{s o m}{%
  \ifstrequal{#3}{Main Results on LiveMCPBench}{\Needspace{15\baselineskip}}{}%
  \IfBooleanTF{#1}{\AnyActOriginalSubsection*{#3}}{%
    \IfNoValueTF{#2}{\AnyActOriginalSubsection{#3}}{\AnyActOriginalSubsection[#2]{#3}}}%
}

\hypersetup{
  hidelinks,
  pdftitle={AnyAct: Universal Action for Self-Evolving Agents},
  pdfauthor={Lingrui Xu, Yangqin Jiang, Jiachang Zhang, Xubin Ren, Chao Huang}
}

\newcommand{\ie}{\textit{i}.\textit{e}.\xspace}
\newcommand{\eg}{\textit{e}.\textit{g}.\xspace}

\theoremstyle{plain}

\theoremstyle{definition}

\theoremstyle{remark}

\def\model{AnyAct\xspace}

\definecolor{blue}{HTML}{4F8AAD}
\definecolor{pink}{HTML}{EDE5FB}
\definecolor{green}{HTML}{C79AB4}
\definecolor{grey}{HTML}{6C5B7B}

\title{AnyAct: Universal Action \\for Self-Evolving Agents}

\author{
  \normalfont % Reset the template's first-row bold; only names are bold below.
  \textbf{Lingrui Xu}\enspace
  \textbf{Yangqin Jiang}\enspace
  \textbf{Jiachang Zhang}\enspace
  \textbf{Xubin Ren}\enspace
  \textbf{Chao Huang}\thanks{Corresponding author.} \\[4pt]
  \normalfont The University of Hong Kong \\[3pt]
  {\small\texttt{\{\href{mailto:lingruixu.db@gmail.com}{lingruixu.db},
  \href{mailto:mrjiangyq99@gmail.com}{mrjiangyq99},
  \href{mailto:xubinrencs@gmail.com}{xubinrencs},
  \href{mailto:chaohuang75@gmail.com}{chaohuang75}\}@gmail.com}} \\[3pt]
  {\small\texttt{\href{mailto:jiachang@connect.hku.hk}{jiachang@connect.hku.hk}}} \\[3pt]
  \faGithub\ \textbf{GitHub Repo:}\enspace
  \href{https://github.com/HKUDS/AnyTool}{\textcolor[rgb]{0,0,1}{\texttt{https://github.com/HKUDS/AnyTool}}}
}

\iclrfinalcopy
\begin{document}

\maketitle
\lhead{Preprint.}

\begin{abstract}
As large language models (LLMs) advance, AI agents are increasingly deployed in open-world environments to tackle complex sequential tasks (\textit{e.g.}, document processing, cross-application collaboration), relying heavily on actions ranging from GUI operations to semantic APIs. However, three core challenges persist: the "scale dilemma" of massive tool ecosystems exceeding LLM context windows, the "non-stationarity" of tool quality due to updates or outages, and the "heterogeneity" of feedback formats (pixels, text, structured data) creating information silos. To address these, we propose \textbf{\model}, a universal action layer that unifies available capabilities into a self-evolving action space, enabling agents to operate efficiently and reliably in large-scale, dynamic tool ecosystems. \model's core design focuses on two objectives: constructing this action space via hierarchical progressive retrieval (filtering task-relevant actions) and test-time reliability evolution (pruning unreliable actions), and enabling reliability-aware action orchestration through a heterogeneous observation grounding module that unifies multi-modal feedback. Additionally, it defines a hybrid action space (primitive + semantic actions) and optimizes for a balance between task success rate and execution cost. Evaluations on LiveMCPBench and OSMCP (a new benchmark we developed for multi-granularity action collaboration) demonstrate state-of-the-art performance. \model delivers substantial performance gains over baseline methods across various LLM base models on LiveMCPBench and improvements are particularly notable for models with constrained native capabilities. On OSMCP, it achieves 77.27\% overall success with only 50 steps, which is half the steps required by most competitors.
\end{abstract}

\section{Introduction}
\label{sec:intro}

As large language model (LLM) capabilities continue to advance~\citep{tran2025multi, fang2025comprehensive, jiang2026openphone}, particularly in reasoning and tool utilization~\citep{guo2025deepseek, li2025search}, AI agents have progressively moved from domain-specific applications~\citep{sun2025genesis, yang2024swe} into open-world environments~\citep{xie2024osworld, yang2025agentic}, taking on complex sequential tasks such as document processing, cross-application collaboration, and data analysis. The successful completion of these tasks depends heavily on the agent's action space, ranging from pixel-level GUI interactions and command-line instructions to semantic APIs that encapsulate complex logic, and the richness of the tool ecosystem~\citep{gan2025rag, li2023api, lu2025build} extends the agents' capability boundaries. In particular, with the widespread adoption of open protocols such as the Model Context Protocol (MCP)~\citep{mcp}, the number of available tools has grown explosively, creating a large ecosystem that spans multiple domains and granularities and providing agents with more efficient pathways for task execution.

However, agents acting in the open world still face three core challenges. First, the tool ecosystem's \textbf{``scale dilemma''}: the full set of API specifications often exceeds an LLM's context window capacity, and exposing all tools directly leads to inefficient retrieval and unclear task relevance~\citep{shi2025aime, lin2025se}. Second, the \textbf{``non-stationarity''} of tool quality: in open environments tools may fail or degrade due to version updates, service outages, or misleading documentation, and static tool selection strategies cannot cope with such dynamic reliability changes. Third, the \textbf{``heterogeneity''} of feedback formats: GUI actions produce pixels, command-line operations return text, and APIs yield structured data, and these modality differences make it difficult for agents to perform cross-tool collaborative reasoning, creating information silos. Existing methods either rely on fixed action sets and therefore lack generalization, or lack dynamic awareness of action reliability and therefore lack robustness; in neither case can the agent evolve its own action space from experience, making it difficult to achieve an optimal tradeoff between task success rate and execution efficiency.

To address the aforementioned challenges, we propose \textbf{\model}, a universal action layer for self-evolving agents. It is designed to enable agents to efficiently and reliably accomplish open-world tasks within large-scale, dynamically evolving tool ecosystems.
% To this end, it unifies GUI primitives, shell commands, and MCP tools into a single action space, and evolves this space from execution feedback instead of retraining the LLM.
The core design of \model focuses on two primary objectives. \textbf{Objective 1:} Constructing a self-evolving action space capable of precisely selecting task-relevant subsets from a massive tool repository while dynamically pruning unreliable tools; \textbf{Objective 2:} Implementing reliability-aware action orchestration that unifies heterogeneous tool feedback and coordinates tools of varying granularities for collaborative execution.

Specifically, \model employs a hierarchical and progressive retrieval mechanism that utilizes the natural structure of the tool ecosystem to perform efficient coarse-grained filtering followed by fine-grained ranking. This enables the construction of a compact yet comprehensive action space during task initialization. Through a test-time reliability evolution mechanism, \model dynamically updates reliability scores based on execution feedback, thereby guiding the agent to prioritize high-quality actions. A heterogeneous observation grounding module integrates feedback from multiple modalities into a unified knowledge space, eliminating information barriers across different actions. Furthermore, \model defines a hybrid action space consisting of both primitive actions and task-relevant semantic actions. It formulates an optimization objective that balances task success rate with execution cost, achieving coordinated improvements in both robustness and efficiency.

\model demonstrates strong performance on both the LiveMCPBench and OSMCP benchmarks. OSMCP is constructed by extending OSWorld~\citep{xie2024osworld} and integrates a rich set of MCP servers that expose agent-native action interfaces, providing realistic support for evaluating multi-granularity action collaboration. Our main contributions can be summarized as follows:
% \vspace{-0.1in}
\begin{itemize}[leftmargin=*]
    \item \textbf{Universal Action Layer for Self-Evolving Agents.} We introduce the first universal action layer that unifies GUI, shell, and MCP actions and supports large-scale dynamic tool ecosystems, enabling self-evolving construction and maintenance of the action space. We also develop the OSMCP benchmark and its accompanying MCP toolset, filling an evaluation gap for multi-granularity action collaboration in open-world tasks.

    \item \textbf{Reliability-Aware Action Orchestration Strategy.} We design a reliability-aware action orchestration strategy that combines test-time evolution of action reliability with a primitive action fallback mechanism, efficiently adapting to challenges such as tool quality fluctuation and service outages in open environments.

    \item \textbf{Cross-Modal Observation Grounding Mechanism.} We develop a cross-modal observation grounding mechanism that maps GUI pixels, command-line text, and structured API feedback into a shared knowledge space, breaking heterogeneous information barriers and supporting collaborative reasoning across different types of actions.
\end{itemize}

\section{Preliminary}
\label{sec:preliminary}
We consider an agent performing a task $I$ through sequential interaction with an open-world environment. At each step $t$, the agent selects an action $a_t \in \mathcal{A}$ based on its observation $o_t$. The overall effectiveness is fundamentally driven by the co-design of the action hierarchy and the orchestration policy:
% \vspace{-0.1in}
\begin{itemize}[leftmargin=*]

    \item \textbf{Action Space.}
    We define a unified action space $\mathcal{A} = \mathcal{A}_{\mathrm{gui}} \cup \mathcal{A}_{\mathrm{cmd}} \cup \mathcal{A}_{\mathrm{api}}$. Specifically, $\mathcal{A}_{\mathrm{gui}}$ provides atomic, pixel-level control for universal applicability. In contrast, $\mathcal{A}_{\mathrm{cmd}}$ and $\mathcal{A}_{\mathrm{api}}$ consist of semantic actions, which encapsulate high-level logic to perform complex tasks efficiently. The effective action space is further shaped by two practical considerations: (i) \textbf{Context-Aware Scaling:} Since the exhaustive library of API specifications is often too vast to be included in an LLM's context window, we retrieve a task-relevant subset $\mathcal{H}(I) \subseteq \mathcal{A}_{\mathrm{api}}$ with $|\mathcal{H}(I)| \leq K$. This yields an instance-specific space $\mathcal{A}_I = \mathcal{A}_{\mathrm{base}} \cup \mathcal{H}(I)$, where $\mathcal{A}_{\mathrm{base}} = \mathcal{A}_{\mathrm{gui}} \cup \mathcal{A}_{\mathrm{cmd}}$ represents the invariant set of fundamental actions available across all tasks. (ii) \textbf{Execution Reliability: }Unlike low-level GUI steps, semantic actions may fail during execution due to environmental shifts or API instabilities. To account for this, we maintain a reliability score $r^{(n)}(a) \in [0,1]$ that evolves across episodes to track the usability of each action.

    \item \textbf{Action Orchestration.}
    Different action types produce heterogeneous feedback, ranging from screenshots ($\mathcal{A}_{\mathrm{gui}}$) and terminal logs ($\mathcal{A}_{\mathrm{cmd}}$) to structured responses ($\mathcal{A}_{\mathrm{api}}$). We aggregate these into a recursive context $c_{t+1} = \mathrm{agg}(c_t, \phi(o_t))$ to guide subsequent decisions. The policy $\pi(a_t \mid I, c_t, \mathcal{A}_I, r^{(n)})$ implements a reliability-aware selection strategy: reliability influences the retrieved action priorities, while the LLM selects among available semantic and base actions using task context and execution feedback.
\end{itemize}

\textbf{Objective.}
% The agent jointly optimizes action space (via $\mathcal{H}$ and $r$) and orchestration (via $\pi$):
% \begin{equation}
%     \max_{\pi, \mathcal{H}, r} \; p(\text{success} \mid I) - \lambda |\tau|,
%     \label{eq:objective}
% \end{equation}
% maximizing success rate while minimizing trajectory $|\tau|$.
The agent aims to jointly optimize the Action Hierarchy $\mathcal{H}$, the Reliability Estimator $r$, and the Orchestration Policy $\pi$. This joint optimization is formulated as:
\begin{equation}\label{eq:objective}
\max_{\pi, \mathcal{H}, r} \; \mathbb{P}(\text{success} \mid I) - \lambda |\tau|,
\end{equation}
where $|\tau|$ denotes the trajectory length. This objective maximizes the success rate while penalizing execution cost via $\lambda$. Specifically, the penalty encourages the policy to leverage high-level actions in $\mathcal{H}$ to reduce steps, while $r$ ensures these actions are selected only when reliable. This formulation achieves an optimal trade-off between operational robustness and efficiency.

\section{The \model\ Framework}
\label{sec:method}

\begin{figure}[t]
    \centering
    \includegraphics[width=1.0\textwidth]{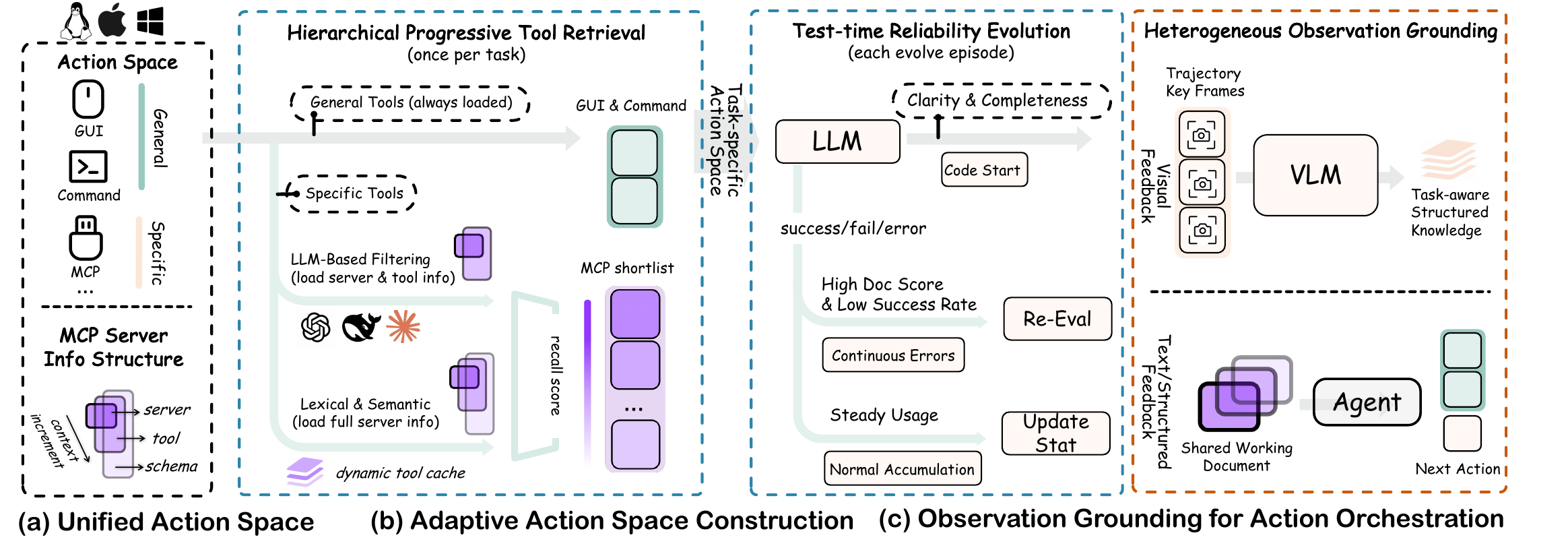}
    \vspace{-0.2in}
    \caption{The overall framework of \model\ for self-evolving agentic automation over a universal action space.}
    \vspace{-0.2in}
    \label{fig:framework}
\end{figure}

% We present \model, a tool-augmented framework for scalable, reliable, and efficient agent automation in open-world environments. Our approach addresses two fundamental challenges: (1) action space construction, which curates task-relevant tools from massive, evolving ecosystems and updates their test-time reliability through execution, and (2) action orchestration, which coordinates heterogeneous action types to maximize task success rate with minimal steps. In the following sections, we detail these components and their integration into a unified automation framework.

We present \model, a universal action layer for scalable, reliable, and efficient self-evolving agent automation in open-world environments.
The central goal of \model\ is to enable an agent to operate in large evolving tool ecosystems while maintaining high task success rates and low execution cost.
To this end, the framework addresses two tightly coupled challenges:
(1) \emph{adaptive action space construction}, which selects and maintains a task-relevant subset of actions from a massive and non-stationary action space and evolves it from execution feedback, and
(2) \emph{reliability-aware action orchestration}, which coordinates heterogeneous actions to complete tasks efficiently and robustly.
In the following, we describe how these components are designed and integrated into a unified framework.

\subsection{Adaptive Action Space Construction}

The full action space $\mathcal{A}$, comprising low-level GUI primitives, shell commands, and high-level semantic APIs, provides complete coverage for open-world tasks in principle. In open tool ecosystems, however, directly exposing $\mathcal{A}$ to the agent is infeasible: $\mathcal{A}$ is both \emph{too large} to fit within the context window and \emph{non-stationary} at test time, as tools may be added, updated, or silently fail. Accordingly, the agent is exposed to an \emph{evolving}, task-conditioned subset $\mathcal{A}_I^{(n)} \subset \mathcal{A}$, constructed per task $I$ and revised across episodes $n$, while retaining low-level primitives as a reliable fallback. Under these practical constraints, such an evolving action space is unavoidable, and it gives rise to two challenges. (i) \textbf{Retrieval efficiency versus coverage:} refreshing $\mathcal{A}_I^{(n)}$ at every step improves precision but incurs trajectory-length overhead, whereas one-shot retrieval is more efficient but may exclude task-critical actions. (ii) \textbf{Test-time reliability:} even retrieved actions may fail or drift during execution, requiring $\mathcal{A}_I^{(n)}$ to be re-ranked and revised based on execution feedback rather than static curation.

To address these challenges, we introduce two complementary mechanisms.
\textbf{Hierarchical Progressive Retrieval} exploits the natural hierarchy of tool ecosystems to enable coarse-to-fine filtering, allowing single-shot retrieval at task initialization while preserving functional coverage.
\textbf{Test-Time Reliability Evolution} tracks action usability through execution outcomes and demotes unreliable actions across episodes.
Together, these mechanisms yield an action space that is both context-efficient and empirically dependable.

\subsubsection{Hierarchical Progressive Retrieval}

% 检索目标是谁，为什么，这里说明我们只检索一次
% Retrieval constructs $\mathcal{H}(I) \subseteq \mathcal{A}_{\mathrm{api}}$ under the budget $|\mathcal{H}(I)| \leq K$. We retrieve only semantic actions in $\mathcal{A}_{\mathrm{api}}$, while keeping base actions $\mathcal{A}_{\mathrm{base}}$ always loaded. This asymmetry provides (1) fallback completeness, $\mathcal{A}_{\mathrm{base}}$ can solve any task via primitives when semantic tools are missing or fail, and (2) single-shot retrieval at task initialization. General-purpose operations (e.g., navigation and file manipulation) are essential but difficult to retrieve via semantic matching against the task description and would otherwise require step-wise retrieval. The effective action space is $\mathcal{A}_I = \mathcal{A}_{\mathrm{base}} \cup \mathcal{H}(I)$.

Our goal is to construct a compact set of task-relevant semantic actions $\mathcal{H}(I)\subseteq\mathcal{A}_{\mathrm{api}}$ under a strict budget $|\mathcal{H}(I)|\le K$, while avoiding repeated retrieval during execution. Accordingly, we retrieve only semantic actions in $\mathcal{A}_{\mathrm{api}}$, and keep base actions $\mathcal{A}_{\mathrm{base}}$ always available. This asymmetric design provides two benefits: (i) \emph{fallback completeness}, as $\mathcal{A}_{\mathrm{base}}$ can solve any task via low-level primitives when semantic tools are missing or fail; and (ii) \emph{single-shot retrieval}, allowing the action space to be constructed once at task initialization. The resulting action space is $\mathcal{A}_I=\mathcal{A}_{\mathrm{base}}\cup\mathcal{H}(I)$.

% 检索的challenge：太大，引出渐进式加载
% The remaining challenge is to construct $\mathcal{H}(I)$ efficiently from thousands of candidates. Tool ecosystems are naturally hierarchical: each server $s \in \mathcal{S}$ hosts tools $\mathcal{A}_s \subseteq \mathcal{A}_{\mathrm{api}}$, and each tool admits a compact metadata view $m(a)=(\mathrm{name},\mathrm{desc})$ distinct from its verbose specification $\mathrm{spec}(a) = (\mathrm{params}, \mathrm{schema}, \mathrm{examples})$. This motivates a progressive disclosure strategy: metadata is loaded upfront for coarse filtering, while full specifications are loaded on-demand only for selected candidates.

Large-scale tool ecosystems are naturally organized into hierarchical registries. At the top level, tools are grouped into providers or servers $s\in\mathcal{S}$ (\eg, MCP servers), each hosting a subset of tools $\mathcal{A}_s\subseteq\mathcal{A}_{\mathrm{api}}$. Each tool further admits a compact metadata view $m(a)=(\mathrm{name},\mathrm{desc})$ that is distinct from its verbose specification $\mathrm{spec}(a)=(\mathrm{params},\mathrm{schema},\mathrm{examples})$. This structure enables a progressive disclosure strategy: metadata can be used for coarse filtering at scale, while full specifications are loaded only for a small set of selected tools.

% 渐进式加载为什么要LLM和sim都有
% \textbf{Coarse-to-Fine Gating.}
% Metadata filtering can be driven by either embedding similarity or LLM reasoning.
% Embedding retrieval is efficient but suffers from an intent-capability mismatch: task instructions express user intent, while tool metadata describes declarative functions, and implicit requirements often emerge only after decomposition.
% LLM reasoning can bridge this gap, but scoring thousands of tools is prohibitively expensive.
% We therefore exploit the registry hierarchy to combine their strengths: we use LLM reasoning only at the server level ($|\mathcal{S}|\!\ll\!|\mathcal{A}_{\mathrm{api}}|$) to obtain a coarse shortlist, and perform fine-grained tool ranking with embeddings within the shortlisted servers.
\textbf{Coarse-to-Fine Gating.}
We adopt a hierarchical retrieval strategy that combines LLM reasoning with embedding-based ranking.
Embedding retrieval scales efficiently but suffers from an intent--capability mismatch: task instructions express user intent, whereas tool metadata describes declarative functions, and implicit requirements often emerge only after task decomposition.
LLM reasoning can bridge this gap, but directly scoring thousands of tools is prohibitively expensive.
To resolve this tension, we exploit the registry hierarchy by applying LLM reasoning only at the top level
($|\mathcal{S}|\!\ll\!|\mathcal{A}_{\mathrm{api}}|$),
followed by embedding-based ranking within the shortlisted servers.

% Formally, let $\mathcal{S}_{\mathrm{plan}}(I)\subseteq\mathcal{S}$ denote the server shortlist returned by the planning stage, and let $\mathcal{H}_{\mathrm{rank}}(I,\mathcal{C})$ return the Top-$K$ tools from candidate pool $\mathcal{C}$. Let $\mathcal{C}_0=\mathcal{A}_{\mathrm{api}}$ be the initial pool (tool metadata) available at retrieval time. When $|\mathcal{C}_0|>N_{\mathrm{thresh}}$, we set $\mathcal{C}=\bigcup_{s\in\mathcal{S}_{\mathrm{plan}}(I)}\mathcal{A}_s$; otherwise we directly rank over $\mathcal{C}_0$.

For task $I$, let $\mathcal{S}_{\mathrm{plan}}(I)\subseteq\mathcal{S}$ denote the coarse server shortlist and $\mathcal{C}_0=\mathcal{A}_{\mathrm{api}}$ the full metadata pool. When $|\mathcal{C}_0|>N_{\mathrm{thresh}}$, the candidate pool is $\mathcal{C}=\bigcup_{s\in\mathcal{S}_{\mathrm{plan}}(I)}\mathcal{A}_s$; otherwise, $\mathcal{C}=\mathcal{C}_0$. Reranking returns $\mathcal{H}(I)=\mathcal{H}_{\mathrm{rank}}(I,\mathcal{C})$.

\begin{itemize}[leftmargin=*]
    \vspace{-0.05in}
    % \item \textbf{Server Shortlisting.}
    % For massive action spaces, we expose the LLM only compact server-level metadata
    % $\{(s,\mathrm{desc}(s),\{m(a)\}_{a\in\mathcal{A}_s})\}_{s\in\mathcal{S}}$,
    % i.e., server descriptions and tool names/descriptions, while withholding verbose specifications.
    % Given task $I$, the LLM outputs:
    % (i) a set of high-confidence tools $\mathcal{A}_{\mathrm{hc}}\subset\mathcal{A}_{\mathrm{api}}$ matched by name,
    % (ii) a set of relevant servers $\mathcal{S}_{\mathrm{sel}}\subseteq\mathcal{S}$ whose tools should be explored,
    % and (iii) a brief rationale $J$ (not used for downstream decisions).
    % We then form the candidate pool
    % \begin{equation}
    % \mathcal{C}_{\mathrm{plan}}
    % =\mathcal{A}_{\mathrm{hc}} \ \cup\ \bigcup_{s\in\mathcal{S}_{\mathrm{sel}}}\mathcal{A}_s.
    % \end{equation}
    % If $|\mathcal{C}_{\mathrm{plan}}|\le K$, we set $\mathcal{H}(I)=\mathcal{C}_{\mathrm{plan}}$; otherwise we pass $\mathcal{C}_{\mathrm{plan}}$ to tool reranking.
    % \vspace{-0.1in}
    \item \textbf{Server Shortlisting.}
    For very large action spaces, we first expose the LLM only to compact server-level metadata
    $\{(s,\mathrm{desc}(s),\{m(a)\}_{a\in\mathcal{A}_s})\}_{s\in\mathcal{S}}$ (\ie, server descriptions and tool names/descriptions). Given a task $I$, the LLM identifies (i) a set of high-confidence tools $\mathcal{A}_{\mathrm{hc}}\subset\mathcal{A}_{\mathrm{api}}$ matched by name, and (ii) a set of relevant servers $\mathcal{S}_{\mathrm{sel}}\subseteq\mathcal{S}$ whose tools should be explored. We then form the candidate pool as follows:
    \begin{equation}
        \mathcal{C}_{\mathrm{plan}} = \mathcal{A}_{\mathrm{hc}} \cup \bigcup_{s \in \mathcal{S}_{\mathrm{sel}}} \mathcal{A}_s.
    \end{equation}
    If $|\mathcal{C}_{\mathrm{plan}}| \le K$, we return $\mathcal{H}(I) = \mathcal{C}_{\mathrm{plan}}$ directly; otherwise, we rerank the candidates.

    % \item \textbf{Tool Reranking.}
    % For moderate action spaces or overflow candidates, we expand the query once: an LLM converts $I$ into capability-oriented keywords $q$ (optionally with a brief rationale),
    % bridging the intent--capability mismatch without scoring each tool with the LLM.
    % We use a hybrid dense--sparse relevance score combining embedding similarity and BM25:
    % \begin{equation}
    % u(a;q)=\alpha\,\langle \mathrm{enc}(q),\mathrm{enc}(m(a))\rangle+(1-\alpha)\,\mathrm{BM25}(q,m(a)).
    % \label{eq:hybrid_score}
    % \end{equation}
    % Here $\mathrm{enc}(\cdot)$ is a dense encoder that maps text to a normalized embedding vector, and $\mathrm{BM25}(\cdot,\cdot)$ is a sparse lexical relevance score~\citep{bm25}.

    \item \textbf{Tool Reranking.}
    When the candidate pool exceeds $K$, the LLM converts task $I$ into capability-oriented keywords $q$. BM25~\citep{bm25} first selects a lexical candidate pool $\mathcal{C}_{\mathrm{lex}}\subseteq\mathcal{C}$, which is then reranked by semantic similarity:
    \begin{equation}\label{eq:hybrid_score}
        u(a;q) = \text{sim}_{\text{sem}}(q,a),
        \qquad a\in\mathcal{C}_{\mathrm{lex}},
    \end{equation}
    where $\text{sim}_{\text{sem}}$ measures similarity between normalized dense embeddings. Lexical filtering narrows the pool, and semantic reranking orders the retained candidates.

\end{itemize}

% Finally, letting $q$ denote the expanded query, we return
% \begin{equation}
% \mathcal{H}(I)=\mathrm{Top}\!-\!K\big(\mathcal{C};\,u(\cdot;q)\big),
% \label{eq:topk_return}
% \end{equation}
% where the candidate pool is $\mathcal{C}=\mathcal{A}_{\mathrm{api}}$ when we skip server shortlisting and $\mathcal{C}=\mathcal{C}_{\mathrm{plan}}$ when reranking overflow candidates from server shortlisting.

After reranking, up to $K$ tools are returned as
\begin{equation}\label{eq:topk_return}
    \mathcal{H}(I) = \mathrm{Top\!-\!K}\big(\{a \in \mathcal{C}_{\mathrm{lex}}\}; u(a;q)\big),
\end{equation}
where $\mathcal{C}_{\mathrm{lex}}$ is the BM25-filtered subset of $\mathcal{C}$. Here, $\mathcal{C}$ is $\mathcal{A}_{\mathrm{api}}$ when server shortlisting is skipped and $\mathcal{C}_{\mathrm{plan}}$ otherwise.

\subsubsection{Test-Time Reliability Evolution}

% 为什么要记录runtime质量，怎么影响retrieve分数
% Progressive retrieval determines \emph{what} semantic tools to load into context; reliability evolution determines \emph{whether} they remain usable at runtime.
% Since open tool ecosystems are non-stationary (e.g., outages, breaking updates, malformed schemas, and misleading documentation), we maintain an execution-driven reliability score $r^{(n)}(a)\in[r_{\min},1]$ for each semantic action $a\in\mathcal{A}_{\mathrm{api}}$, updated across episodes $n$.
% For any candidate action with base retrieval score $u^{(n)}(a)$, we fuse reliability multiplicatively:
% \begin{equation}
% u^{(n)}(a)\ \leftarrow\ u^{(n)}(a)\cdot r^{(n)}(a),
% \label{eq:score_fusion}
% \end{equation}
% so persistently failing tools are demoted in ranking while remaining recoverable when they become healthy again.

While the above process determines \emph{which} semantic actions to load, runtime reliability decides \emph{whether} these actions remain usable during execution. Open tool ecosystems are non-stationary: tools can fail due to outages, breaking updates, malformed schemas, or misleading documentation. To account for this, we maintain a per-action reliability score $r^{(n)}(a)\in[r_{\min},1]$, updated across episodes $n$, which acts as a multiplicative gate on the retrieval score:
\begin{equation}\label{eq:score_fusion}
    u^{(n)}(a) \ \leftarrow\ u^{(n)}(a)\cdot r^{(n)}(a).
\end{equation}
This ensures that persistently failing actions are gradually demoted, while still allowing recovery if subsequent executions succeed.
% This is what makes the action space \emph{self-evolving}: the agent revises its own action interface purely from execution experience, without any parameter updates to the underlying LLM.

\textbf{Execution Profiling.}
% We maintain a quality record for each semantic action $a$, indexed by a unique endpoint key
% $\kappa(a)=\langle \mathrm{server}(a),\mathrm{tool}(a)\rangle$.
% For each $\kappa(a)$, we store a sliding window $\mathcal{Q}_a^{(n)}$ of the $M$ most recent executions (we use $M{=}100$), with each record
% \begin{equation}
% e_i=\langle \texttt{success}_i,\ \texttt{latency}_i,\ \texttt{error\_type}_i\rangle.
% \label{eq:exec_record}
% \end{equation}
% From $\mathcal{Q}_a^{(n)}$ we derive the windowed success rate
% \begin{equation}
% R_a^{(n)}=\frac{1}{|\mathcal{Q}_a^{(n)}|}\sum_{e\in\mathcal{Q}_a^{(n)}} \mathbf{1}\{\texttt{success}(e)\},
% \label{eq:success_rate}
% \end{equation}
% and the trailing consecutive failure count $C_a^{(n)}$.
% Let $N_a^{(n)}$ denote the cumulative invocation count of $a$ through episode $n$.
% The sliding window balances responsiveness to behavioral shifts against statistical stability.
For each tool $a$, we maintain a recent execution history to capture runtime behavior, from which we compute two key statistics: the recent success rate $R_a^{(n)}$, summarizing stable performance, and the trailing consecutive failure count $C_a^{(n)}$, which quickly detects abrupt breakages. These statistics are tracked using a sliding window over recent executions, without exposing low-level details such as individual latency or error codes.
Here, success denotes the runtime execution-status flag returned by the action
backend; task completion is assessed separately by the benchmark evaluator.

\textbf{Reliability Update.}
% We implement tool reliability as a single bounded update rule $r^{(n)}(a)\in[r_{\min},1]$ that is carried across episodes $n$ and used as a multiplicative gate on retrieval scores (Eq.~\ref{eq:score_fusion}).
% The update is computed from two complementary execution statistics obtained in profiling: the recent success rate $R_a^{(n)}$ (for stable calibration under noisy outcomes) and the trailing consecutive failure count $C_a^{(n)}$ (for rapid detection of abrupt breakages).
We update $r^{(n)}(a)$ from the recent success rate $R_a^{(n)}$ and consecutive failure count $C_a^{(n)}$ as follows:

\begin{itemize}[leftmargin=*]
% \vspace{-0.05in}
    \item \textbf{Cold start.}
    % If a tool has insufficient evidence, $N_a^{(n)}<N_{\min}$, we set $r^{(n)}(a)=1$ to avoid premature suppression from a few early, atypical executions.
    % Here $N_a^{(n)}$ is the cumulative invocation count of tool $a$ across episodes, which prevents rarely used tools from remaining permanently in a cold regime and reduces bias from incidental failures during initial exploration.
    Let $N_a^{(n)}$ denote the cumulative invocation count. For tools with limited history ($N_a^{(n)} < N_{\min}$), we set $r^{(n)}(a) = 1$ to avoid penalizing tools prematurely. This allows the system to collect sufficient evidence before demoting a tool.

    % \vspace{-0.1in}
    \item \textbf{Update rules.}
    % Otherwise, we compute a bounded penalty from rate calibration and streak suppression:
    % \begin{equation}
    % r^{(n)}(a)=\mathrm{clip}\!\Big(r_{\mathrm{base}}(R_a^{(n)})-\delta_{\mathrm{consec}}(C_a^{(n)}),\ r_{\min},\ 1\Big),
    % \label{eq:r_update}
    % \end{equation}
    % where $[x]_+=\max(x,0)$ and $\mathrm{clip}(x,l,u)=\min(u,\max(l,x))$, and
    % \begin{equation}
    % \begin{aligned}
    % r_{\mathrm{base}}(R) &= 1-(1-r_{\min})\Big[1-\tfrac{R}{\theta_\rho}\Big]_+,\\
    % \delta_{\mathrm{consec}}(C) &= \min\!\big(\delta_{\max},\ \delta_{\text{step}}\,[C-F]_+\big).
    % \end{aligned}
    % \label{eq:base_and_consec}
    % \end{equation}
    % The base term $r_{\mathrm{base}}$ defines a pass region ($R\ge\theta_\rho$) and smoothly downweights tools below threshold, avoiding unstable rank flips due to small estimation noise around $\theta_\rho$.
    % The streak term $\delta_{\mathrm{consec}}$ acts as a circuit breaker: once failures persist beyond patience $F$, it rapidly demotes likely-broken endpoints regardless of historical averages.
    After cold start, we subtract the consecutive-failure penalty from the base reliability and clip the result to $[r_{\min},1]$:

    \begin{equation}\label{eq:r_update}
        r^{(n)}(a) = \mathrm{clip}\Big(\underbrace{r_{\mathrm{base}}(R_a^{(n)})}_{\text{stable performance}}
        - \underbrace{\delta_{\mathrm{streak}}(C_a^{(n)})}_{\text{circuit breaker}}, \ r_{\min}, 1\Big),
    \end{equation}

    where $\mathrm{clip}(x,l,u)=\min(u,\max(l,x))$. The base reliability downweights tools with low success rates:
    \begin{equation}\label{eq:base_and_consec}
        r_{\mathrm{base}}(R) =
        \begin{cases}
            1, & R \ge \theta_\rho, \\
            1-(1-r_{\min}) \frac{\theta_\rho - R}{\theta_\rho}, & R < \theta_\rho,
        \end{cases}
    \end{equation}

    and the streak penalty rapidly demotes tools with consecutive failures beyond a patience threshold $F$, with $[x]_+=\max(x,0)$:
    \begin{equation}
        \delta_{\mathrm{streak}}(C) = \min\big(\delta_{\max}, \delta_{\text{step}} \cdot [C-F]_+\big).
    \end{equation}

    \item \textbf{Behavioral impact.}
    % The update induces three regimes that reshape the retrieved action interface via Eq.~\ref{eq:score_fusion}:
    % (i) an \emph{exploration} regime when $N_a^{(n)}<N_{\min}$, where tools are not penalized and the system can accumulate evidence;
    % (ii) a \emph{stable} regime when $R_a^{(n)}\ge\theta_\rho$ and $C_a^{(n)}\le F$, where reliable tools retain high rank and are preferred for efficient semantic execution; and
    % (iii) a \emph{failure} regime, where $R_a^{(n)}<\theta_\rho$ triggers gradual demotion while $C_a^{(n)}>F$ activates the circuit breaker and rapidly suppresses likely-broken endpoints, shifting selection toward healthier alternatives or fallback primitives.
    % Clipping guarantees $r^{(n)}(a)\ge r_{\min}$, so tools are never permanently excluded and can recover in ranking once subsequent executions succeed.
    The update creates three regimes:
    (i) \emph{Exploration}, for tools with insufficient history, where no penalty is applied;
    (ii) \emph{Stable}, where tools with high success rates and few consecutive failures retain high rank, promoting efficient semantic execution;
    (iii) \emph{Failure}, where low success or persistent failures gradually or rapidly demote tools, shifting selection toward reliable alternatives or fallback primitives.
    Clipping ensures $r^{(n)}(a)\ge r_{\min}$, so tools can recover in ranking once their execution improves.
    % \vspace{-0.1in}
\end{itemize}
% \vspace{-0.1in}
% We use fixed defaults in all experiments (e.g., $N_{\min}{=}5$, $\theta_\rho{=}0.4$, $F{=}3$, $\delta_{\text{step}}{=}0.1$, $\delta_{\max}{=}0.3$, $r_{\min}{=}0.2$).

\textbf{Evolutionary Maintenance.}
% To handle ecosystem drift beyond per-episode updates, we perform lightweight maintenance at session boundaries.
% For each tool, we compute a specification fingerprint $h(a)=\mathrm{hash}(\mathrm{spec}(a))$.
% When $h(a)$ changes, we treat the endpoint as a new instance by resetting its execution history $\mathcal{Q}_a$ and re-initializing $r^{(n)}(a)=1$, preventing stale statistics from being carried across interface changes.
% In addition, for tools that remain unreliable over extended use (e.g., $R_a^{(n)}<\theta_\rho$ for many updates), we optionally trigger an LLM-based diagnosis of documentation quality along two dimensions, \emph{clarity} and \emph{completeness}.
% The diagnosis is used to flag tools for offline repair (e.g., rewriting descriptions or fixing schemas), closing the loop between runtime feedback and tool-catalog quality.
To handle long-term ecosystem drift, we perform lightweight maintenance at session boundaries. Specifically, we compute a specification fingerprint $h(a)$ for each tool, capturing its core definition and usage requirements. When the specification changes, we refresh the cached documentation assessment while retaining the execution history. Tools that remain unreliable over extended periods can optionally be flagged for offline diagnosis or repair (\eg, to improve documentation clarity and completeness).

\subsection{Heterogeneous Observation Grounding}

Agents acting over a universal action space receive feedback in multiple forms: visual observations from GUI interactions (screenshots), text output from shell commands, and structured responses from semantic APIs. While text and structured outputs can be directly consumed by reasoning modules, visual observations pose a challenge: it is often unclear whether an action succeeded and how to extract actionable information for subsequent steps.

\textbf{Grounding Operator.}
To unify these heterogeneous signals, we introduce a grounding operator $\phi(\cdot)$ that converts all observations into a shared knowledge representation $\mathcal{K}$. For text and structured outputs, grounding applies lightweight normalization, such as truncating long outputs or linearizing structured data into compact textual records. The main challenge lies in visual observations, which require converting raw pixels into structured knowledge that downstream reasoning can use.

\textbf{Visual Grounding.}
We use a vision-language model (VLM) to extract actionable elements from screenshots, such as text, numbers, tables, and identifiers like URLs, file paths, or element IDs, while ignoring decorative content. The main idea is to treat screenshots as documents to parse rather than scenes to describe, producing machine-readable knowledge. For sequences with multiple frames, we select representative keyframes to balance coverage and context efficiency.

\textbf{Knowledge-Driven Planning.}
The unified knowledge representation $\phi(o_t) \in \mathcal{K}$ enables reasoning that is independent of the observation modality. Identifiers extracted from APIs can parameterize GUI actions, while states grounded from screenshots can verify the results of executed commands. Grounded observations are accumulated into an episodic context $c_{t+1} = \mathrm{agg}(c_t, \phi(o_t))$, which serves as input to the policy $\pi(a_t \mid I, c_t, \mathcal{A}_I, r^{(n)})$, providing a single, consistent substrate for sequential planning without modality-specific handling.

\section{Evaluation}
\label{sec:evaluation}

% 可以考虑添加对于各个section的介绍

\subsection{Experimental Setup}

% \textbf{Benchmark}. To evaluate \model, we used two benchmarks: LiveMCPBench~\citep{mo2025livemcpbench} and OSWorld~\citep{xie2024osworld}. LiveMCPBench includes 70 MCP servers and 527 tools across six domains, with 95 practical, multi-step tasks solvable via MCP tools. OSWorld covers computer tasks involving real web and desktop applications, OS file I/O, and cross-application workflows.

\textbf{Benchmarks.} We evaluate \model on LiveMCPBench~\citep{mo2025livemcpbench} and OSMCP. LiveMCPBench includes 70 MCP servers and 527 tools across six domains, with 95 practical, multi-step tasks solvable via MCP tools. OSMCP is a computer-use benchmark built on OSWorld~\citep{xie2024osworld}: it selects 198 tasks (including 21 infeasible ones) from seven applications, \ie, LibreOffice Calc, Impress, and Writer, GIMP, Thunderbird, VLC, and VS Code, preserving their instructions and evaluators. For these applications we develop 122 MCP tools (45 Atomic, 69 Compound, and 8 Workflow), so OSMCP currently covers only the OSWorld subset whose software has MCP support. See Appendix~\ref{app:osmcp} for OSMCP statistics and Appendix~\ref{app:osmcp_tools} for the full tool list.

\textbf{Models and Baselines.} For the experiments on LiveMCPBench, we evaluate four frontier models: Claude-Sonnet-4~\citep{claude-4}, DeepSeek-V3~\citep{liu2024deepseek}, DeepSeek-R1~\citep{guo2025deepseek}, and Qwen3-235B~\citep{yang2025qwen3}. Following the setting in LiveMCPBench, we employ DeepSeek-V3 as our evaluation model. For the experiments on OSMCP, we use Claude-Sonnet-4.5 as the base LLM of \model. Moreover, comparisons use two groups of baselines: (1) End-to-end GUI models: Claude-Sonnet-4.5~\citep{claude-sonnet-45}, OpenCUA-72B~\citep{wang2025opencua}, UI-TARS-2~\citep{wang2025ui}, EvoCUA~\citep{xue2026evocua}, DeepMiner-Mano-72B~\citep{fu2025mano}, and Seed-1.8~\citep{seedseed1}. (2) Agentic frameworks: Jedi-7B~\citep{xie2025scaling}, CoACT-1~\citep{song2025coact}, GTA1~\citep{yang2025gta1}, Agentic-Lybic-Maestro~\citep{guo2025agentic}, OS-Symphony~\citep{yang2026symphony}, GBOX Agent~\citep{gbox}, Agent S3~\citep{gonzalez2025unreasonable}, and UiPath Screen Agent~\citep{UIPath}.

\vspace{-0.05in}
\subsection{Main Results on LiveMCPBench}
\vspace{-0.05in}
\begin{table}[t]
    \centering
    \caption{Main Results on LiveMCPBench.}
    \vskip -0.05in
    \small
    \setlength{\tabcolsep}{4pt}
    \resizebox{\textwidth}{!}{
    \begin{tabular}{l|l| c c c c c c |c}
        \toprule
        \textbf{Model} & \textbf{Method} & \textbf{Office} & \textbf{Leisure} & \textbf{Travel} & \textbf{Lifestyle} & \textbf{Finance} & \textbf{Shopping} & \textbf{Overall ($\mathbf{\%}$)}\\
        \midrule
        \multirow{4}{*}{Claude-Sonnet-4}
            & Naive & 90.32 & 64.29 & 75.00 & 80.00 & 78.57 & 66.67 & 78.95 \\
            & \model w/o Retrieval & 80.65 & 71.43 & 75.00 & 93.33 & 90.91  & 77.78 & 81.52\\
            & \model w/o Evolution & 83.87 & 78.57 & 58.33 & 80.00 & 91.67 & 33.33 & 75.27 \\
            \cellcolor{white} & \cellcolor{gray!25} \textbf{\model} & \cellcolor{gray!25} \textbf{93.55} & \cellcolor{gray!25} \textbf{92.86} & \cellcolor{gray!25} \textbf{100.00} & \cellcolor{gray!25} \textbf{100.00} & \cellcolor{gray!25} \textbf{83.33} & \cellcolor{gray!25} \textbf{88.89} & \cellcolor{gray!25} \textbf{93.55} \\
        \midrule
        \multirow{4}{*}{DeepSeek-V3}
            & Naive & 41.94 & 42.86 & 50.00 & 40.00 & 28.57 & 55.56 & 42.11 \\
            & \model w/o Retrieval & 51.61 & 42.86 & 33.33 & 60.00 & 50.00 & 33.33 & 47.31 \\
            & \model w/o Evolution & 70.97 & 50.00 & 50.00 & 66.67 & 50.00 & 33.33 & 58.06 \\
            \cellcolor{white} & \cellcolor{gray!20} \textbf{\model} & \cellcolor{gray!20} \textbf{96.77} & \cellcolor{gray!20} \textbf{78.57} & \cellcolor{gray!20} \textbf{75.00} & \cellcolor{gray!20} \textbf{86.67} & \cellcolor{gray!20} \textbf{83.33} & \cellcolor{gray!20} \textbf{66.67} & \cellcolor{gray!20} \textbf{84.95} \\
        \midrule
        \multirow{4}{*}{DeepSeek-R1}
            & Naive & 41.94 & 50.00 & 58.33 & 46.67 & 50.00 & 55.56 & 48.42 \\
            & \model w/o Retrieval & 70.97 & 57.14 & 58.33 & 60.00 & 75.00 & 44.44 & 63.44 \\
            & \model w/o Evolution & 64.52 & 64.29 & 41.67 & 80.00 & 50.00 & 33.33 & 59.14 \\
            \cellcolor{white} & \cellcolor{gray!20} \textbf{\model} & \cellcolor{gray!20} \textbf{83.87} & \cellcolor{gray!20} \textbf{85.71} & \cellcolor{gray!20} \textbf{66.67} & \cellcolor{gray!20} \textbf{93.33} & \cellcolor{gray!20} \textbf{50.00} & \cellcolor{gray!20} \textbf{77.78} & \cellcolor{gray!20} \textbf{78.49} \\
        \midrule
        \multirow{4}{*}{Qwen3-235B}
            & Naive & 54.84 & 35.71 & 41.67 & 53.33 & 50.00 & 44.44 & 48.42 \\
            & \model w/o Retrieval & 58.06 & 57.14 & 50.00 & 53.33 & 58.33 & 66.67 & 56.99 \\
            & \model w/o Evolution & 74.19 & 57.14 & 66.67 & 60.00 & 58.33 & 66.67 & 65.59 \\
            \cellcolor{white} & \cellcolor{gray!20} \textbf{\model} & \cellcolor{gray!20} \textbf{80.65} & \cellcolor{gray!20} \textbf{78.57} & \cellcolor{gray!20} \textbf{66.67} & \cellcolor{gray!20} \textbf{86.67} & \cellcolor{gray!20} \textbf{75.00} & \cellcolor{gray!20} \textbf{66.67} & \cellcolor{gray!25} \textbf{77.42} \\
        \bottomrule
    \end{tabular}}
    \vspace{-0.1in}
    \label{tab:exp_liveMCP}
\end{table}

\setcounter{topnumber}{1}
\suppressfloats[t]
\begin{table}[t]
\begin{minipage}{\textwidth}
    \centering
    \captionsetup{skip=10pt}
    \caption{Main results on OSMCP. Step budget denotes the maximum number of steps allowed for each system.}
    \vskip 0in
    \scriptsize
        \setlength{\tabcolsep}{0.6mm}
    \resizebox{1\textwidth}{!}{
    \begin{tabular}{l c c| c c c c c c c |c c}
        \toprule
        % \textbf{Method} & \textbf{Model} & \textbf{Steps} & \textbf{vlc} & \textbf{gimp} & \textbf{thunderbird} & \textbf{libreoffice\_calc} & \textbf{libreoffice\_writer} & \textbf{libreoffice\_impress} & \textbf{vs\_code} & Overall (/)& Overall ($\%$)\\
        \textbf{Method} & \textbf{Model} & \textbf{Steps} & \textbf{vlc} & \textbf{gimp} & \textbf{thunderbird} & \textbf{calc} & \textbf{writer} & \textbf{impress} & \textbf{vs\_code} & Overall (/)& Overall ($\%$)\\
        \midrule
        \rowcolor{gray!10}
        \multicolumn{12}{c}{\textit{Foundation E2E GUI}}\\
        Claude-Sonnet-4.5 & --- & 15 & 4.89/17 & 13.00/26 & 10.00/15 & 17.00/47 & 14.00/23 & 22.96/47 & 14.00/23 & 95.85/198 & 48.41 \\
        opencua-72b-preview & --- & 50 & 8.87/17 & 14.00/26 & 12.00/15 & 18.00/47 & 13.00/23 & 26.96/47 & 18.00/23 & 110.83/198 & 55.97 \\
        UI-TARS-2-2509 & --- & 100 & 8.49/17 & 13.00/26 & 11.00/15 & 31.00/47 & 14.00/23 & 26.50/47 & 17.00/23 & 120.99/198 & 61.11 \\
        Claude-Sonnet-4.5 & --- & 50 & 9.00/17 & 15.00/26 & 10.00/15 & 31.00/47 & 15.00/23 & 27.02/47 & 16.00/23 & 123.02/198 & 62.13 \\
        EvoCUA-20260105 & --- & 50 & 8.39/17 & 20.00/26 & 12.00/15 & 26.00/47 & 16.0/23 & 27.96/47 & 20.00/23 & 130.35/198 & 65.83 \\
        DeepMiner-Mano-72B & --- & 100 & 6.00/17 & 23.00/26 & 10.00/15 & 27.00/47 & 18.00/23 & 28.96/47 & 18.00/23 & 130.96/198 & 66.14 \\
        Claude-Sonnet-4.5 & --- & 100 & 9.89/17 & 14.00/26 & 9.00/15 & 34.00/47 & 18.97/23 & 31.96/47 & 17.00/23 & 134.82/198 & 68.09 \\
        Seed-1.8 & --- & 100 & 10.67/17 & 18.00/26 & 12.00/15 & 33.00/47 & 18.88/23 & 28.62/47 & 17.00/23 & 138.17/198 & 69.78 \\
        \midrule
        \rowcolor{gray!10}
        \multicolumn{12}{c}{\textit{Agentic Framework}}\\
        Jedi-7B & OpenAI-o3 & 100 & 9.25/17 & 18.00/26 & 12.00/15 & 20.00/47 & 13.00/23 & 20.96/47 & 19.00/23 & 112.21/198 & 56.67 \\
        CoACT-1 & --- & 150 & 12.23/17 & 17.00/26 & 11.00/15 & 33.00/47 & 17.00/23 & 23.63/47 & 18.00/23 & 131.86/198 & 66.60 \\
        GTA1 & GPT-5 & 100 & 9.84/17 & 20.00/26 & 12.00/15 & 30.00/47 & 13.97/23 & 30.80/47 & 19.00/23 & 135.61/198 & 68.49 \\
        Agentic-Lybic-Maestro& --- & 100 & 13.72/17 & 19.00/26 & 10.00/15 & 28.00/46 & 15.00/23 & 32.63/47 & 19.00/23 & 137.35/197 & 69.72 \\
        OS-Symphony & GPT-5 & 50 & 10.46/17 & 21.00/26 & 12.00/15 & 36.00/47 & 17.97/23 & 22.94/47 & 18.00/23 & 138.37/198 & 69.88 \\
        GBOX Agent & --- & 15 & 7.87/17 & 20.00/26 & 10.00/15 & 35.00/47 & 15.00/22 & 29.96/47 & 21.00/23 & 138.83/197 & 70.47 \\
        Agent S3 (N=1) & Claude-Opus-4.5 & 100 & 12.70/17 & 13.00/26 & 9.00/15 & 40.00/47 & 18.00/23 & 30.99/47 & 16.00/23 & 139.69/198 & 70.55 \\
        UiPath Screen Agent & Claude-Opus-4.5 & 100 & 10.34/17 & 18.00/26 & 11.00/15 & 37.00/47 & 16.00/23 & 32.99/47 & 18.00/23 & 143.33/198 & 72.39 \\
        % Agent S3 (N=10) & Claude-Opus-4.5 + GPT-5 & 100 & 10.70/17 & 19.00/26 & 11.00/15 & 41.00/47 & 19.97/23 & 33.96/47 & 19.00/23 & 154.63/198 & 78.10 \\
        \midrule
        \model w/o GUI & Claude-Sonnet-4.5 & 15 & 11.00/17 & 15.00/26 & 11.00/15 & 30.00/47 & 13.00/23 & 20.00/47 & 17.00/23 & 117.00/198 & 59.09 \\
        \rowcolor{gray!25}
        \textbf{\model} & Claude-Sonnet-4.5 & 50 & 13.00/17 & 22.00/26 & 10.00/15 & 42.00/47 & 18.00/23 & 29.00/47 & 19.00/23 & 153.00/198 & 77.27 \\
        \bottomrule
    \end{tabular}}
    \label{tab:exp_OSWorld}
\end{minipage}

\end{table}

We carry out extensive experiments on LiveMCPBench to evaluate the ability of \model to solve problems via MCP services and to assess the effectiveness of key modules in the proposed method. The corresponding results are summarized in Table~\ref{tab:exp_liveMCP}. Here, \texttt{Naive} refers to the LiveMCPBench baseline built on the ReAct~\citep{yao2022react} framework with the MCP-Zero~\citep{fei2025mcp} retrieval strategy; \texttt{\model w/o Retrieval} indicates a variant where the Hierarchical Progressive Retrieval module of \model is replaced by the MCP-Zero retrieval strategy; and \texttt{\model w/o Evolution} denotes a variant obtained by removing the Test-Time Reliability Evolution module from \model.
% Key findings are as follows:

\textbf{(i) Outstanding proficiency in utilizing MCP APIs and strong generalization.} From the results, \model yields substantial gains over the \texttt{Naive} variant across different LLMs, and for LLMs originally constrained by model capability (\textit{i.e.,} DeepSeek-V3, DeepSeek-R1, and Qwen3-235B) the performance improvements reach nearly 100\%. This indicates that \model imposes relatively low base capability requirements on LLMs: when paired with LLMs of varying strengths, it can effectively enhance their ability to utilize MCP APIs. This improvement is attributable to our retrieval and evolution modules. Hierarchical Progressive Retrieval exploits the natural hierarchy in tool organization for multi-granularity filtering, greatly narrowing and optimizing the action space LLMs must handle; Test-Time Reliability Evolution enables LLMs to continuously assess the quality of different MCP APIs during testing, allowing them to progressively adopt reliable, high-quality APIs and thereby further improve task success rates.

\WFclear
\textbf{(ii) Effective retrieval and evolution module design.} It is clear that both the \texttt{\model w/o Retrieval} and \texttt{\model w/o Evolution} variants suffer substantial performance degradation compared with the full \model. This validates the advantage of our novel Hierarchical Progressive Retrieval over conventional retrieval methods: by applying multi-granularity filtering to the tool set, it accurately identifies the tools required at the start of a task and markedly improves retrieval effectiveness. Meanwhile, Test-Time Reliability Evolution effectively addresses the challenge of heterogeneous tool quality within the tool pool; through continuous evaluation and learning during testing, it steers LLMs toward using high-quality tools.

\subsection{Main Results on OSMCP}

% \vspace{0.2in}

\suppressfloats[t]
\begin{table}[t]
\noindent\begin{minipage}{\textwidth}
\centering
\captionsetup{hypcap=false}
\captionof{table}{Case study: GUI-MCP backend switching for spreadsheet automation in LibreOffice Calc.}
\label{tab:case_study_calc}
\small
\renewcommand{\arraystretch}{0.95}
\setlength{\aboverulesep}{1pt}
\setlength{\belowrulesep}{1pt}
\begin{tabular}{@{}p{\textwidth}@{}}
\toprule
\textbf{Task ID:} \texttt{04d9aeaf-7bed-4024-bedb-e10e6f00eb7f} \\
\midrule
\textbf{Task Instruction:} In a new sheet with 4 headers ``Year'', ``CA changes'', ``FA changes'', and ``OA changes'', calculate the percentage annual changes compared to last year in 2015 to 2019 for the Current Assets, Fixed Assets, and Other Assets columns. \\
\midrule
\textbf{Stepwise Execution (Reasoning $\rightarrow$ Tool Call):} \\[0.2em]
\begin{tabular}{@{}l@{\hspace{0.6em}}c@{\hspace{0.6em}}p{0.74\textwidth}@{}}
\texttt{Step 0 (GUI):} & & ``Check if LibreOffice Calc is open with the balance sheet file ready.'' \\
& $\rightarrow$ & \texttt{gui\_agent(task\_description="Verify Calc open with SmallBalanceSheet.xlsx", max\_steps=15)} \\
\texttt{Step 1 (MCP):} & & ``Let me check the current document to understand the data structure.'' \\
& $\rightarrow$ & \texttt{libreoffice-calc.get\_all\_document\_contents(max\_rows=100, max\_cols=20)} \\
\texttt{Step 2 (MCP):} & & ``Now I'll create a new sheet with the 4 headers.'' \\
& $\rightarrow$ & \texttt{libreoffice-calc.create\_new\_sheet(sheet\_name="Annual Changes")} \\
\texttt{Step 3 (MCP):} & & ``Set up the headers and calculate the percentage changes.'' \\
& $\rightarrow$ & \texttt{libreoffice-calc.set\_cell\_content(cells=[A1:D1 headers, A2:A6 years, B2:D6 formulas])} \\
\texttt{Complete:} & & ``Created sheet `Annual Changes' with percentage annual changes for 2015--2019.'' \\
\end{tabular} \\
\midrule
\textbf{Key Execution Snapshots (cropped worksheet and sheet-tab views):} \\[0.3em]
\begin{minipage}[b]{0.46\textwidth}
\centering
\includegraphics[width=\linewidth]{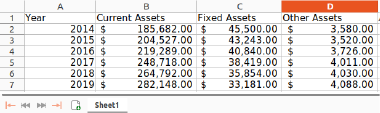}
\par\vspace{2pt}{\normalfont\small (a) Initial state}
\end{minipage}\hfill
\begin{minipage}[b]{0.20\textwidth}
\centering
\includegraphics[width=\linewidth]{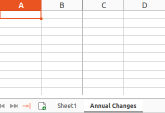}
\par\vspace{2pt}{\normalfont\small (b) Sheet created}
\end{minipage}\hfill
\begin{minipage}[b]{0.31\textwidth}
\centering
\includegraphics[width=\linewidth]{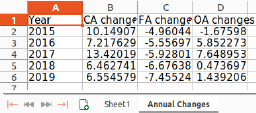}
\par\vspace{2pt}{\normalfont\small (c) Formulas populated}
\end{minipage} \\[0.1em]
\bottomrule
\end{tabular}
\end{minipage}

\end{table}

To more comprehensively evaluate \model's performance on real open-world tasks, we conducted experiments on OSMCP. This benchmark, developed on top of the computer-using benchmark OSWorld, integrates a rich set of MCP servers and can thoroughly assess \model's ability to complete real open-world tasks using tools of different granularity (MCP APIs, shell, GUI). We report the corresponding results in Table~\ref{tab:exp_OSWorld}, where \texttt{\model w/o GUI} denotes the variant that does not use low-level actions (\textit{i.e.}, the GUI action) and runs under a 15-step budget. We evaluate \model under a 50-step budget, counting each GUI action within a subtask separately. Key findings are as follows:

\textbf{(i) Effective and efficient performance on CUA tasks.} According to the results, \model achieves SOTA performance on OSMCP. Moreover, under a setting where other competitive methods require 100 steps, our method is highly efficient, needing only 50 steps. When compared with agentic-framework approaches, we still outperform them despite using relatively weaker LLMs (Claude-4.5-Sonnet vs. Claude-4.5-Opus). This advantage stems from the large number of high-quality MCP servers integrated into OSMCP combined with \model's novel retrieval and evolution designs. By invoking MCP APIs, the model can carry out complex operations directly, substantially reducing task complexity and the number of required steps, thereby delivering simultaneous gains in effectiveness and efficiency.

\textbf{(ii) Low-level actions are indispensable for open-world tasks.} From experiments with the variant \texttt{\model w/o GUI}, we observe markedly worse performance than the full \model. This is because, although high-level actions (\textit{i.e.}, MCP APIs and shell) are very efficient, many real-world open-world tasks depend on fine-grained operations. While fine-grained (\textit{i.e.}, GUI) operations are often less efficient and entail more execution steps, they are more broadly applicable than high-level actions and are indispensable in many scenarios. Therefore, combining high-level and low-level actions to work collaboratively---leveraging the strengths of each---is necessary to complete open-world tasks both effectively and efficiently.

\subsection{Case Study on OSMCP}

To intuitively illustrate how tools at different levels switch and operate within \model on open-world tasks, we conduct a case study on OSMCP, shown in Table~\ref{tab:case_study_calc}.

This example involves creating and computing operations in Office software. We observe that the agent must first verify whether the balance sheet file is already open---a goal that cannot be achieved via the higher-level, integrated MCP APIs. At this point, low-level, highly generalizable GUI actions are needed to intervene, leveraging visual perception to accomplish the check. Subsequently, for operations such as \texttt{Create\_New\_Sheet} and \texttt{Calculate}, APIs provided by MCP servers can accomplish the goals efficiently and accurately, avoiding the tedious, lengthy, and potentially error-prone nature of purely GUI-based actions. Therefore, for CUA tasks, combining GUI actions with MCP APIs is a highly promising solution.

\subsection{In-Depth Analysis}

\label{sec:reliability_analysis}

\begin{wrapfigure}[16]{r}{0.50\textwidth}
    \vspace{\dimexpr-\intextsep+10pt\relax}
    \centering
    \includegraphics[width=\linewidth]{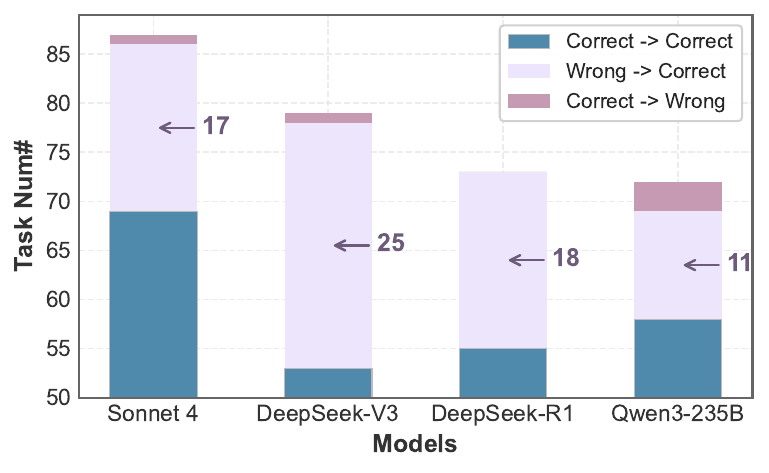}
    \captionsetup{font=normalsize,skip=10pt}
    \caption{Evolution effect across different LLMs on LiveMCPBench. \textcolor{blue}{Blue}: correct before and after evolution; \textcolor{pink!45!black}{pink}: incorrect-to-correct; \textcolor{green!60!black}{purple}: correct-to-incorrect; \textcolor{grey}{deep-purple}: net gain.}
    \label{fig:figure_evolution}

\end{wrapfigure}

To comprehensively evaluate the effect of Test-Time Reliability Evolution, we compile statistics on how evolution influences different LLMs on LiveMCPBench; the corresponding results are shown in Figure~\ref{fig:figure_evolution}. In addition, we select representative examples for case studies.

\Needspace{18\baselineskip}

\textbf{Effect of reliability evolution.} Figure~\ref{fig:figure_evolution} shows that the evolution procedure yields substantial overall gains across different LLMs, validating the effectiveness of our method. For Claude-Sonnet-4, DeepSeek-V3, and Qwen3-235B, however, some tasks that were previously solvable fail after evolution. A likely reason is that the evolution module demotes tools that are unstable or have low success rates when no substitutes exist; tasks that had succeeded only by relying on such tools then fail because those tools are no longer retrieved. This underscores an inherent trade-off between pruning unreliable tools and preserving tools that are uniquely indispensable for certain tasks.

\textbf{The Showing Case on LiveMCPBench.}
We illustrate how self-evolution improves tool selection with a recurring failure in LiveMCPBench.
The tool \texttt{duckduckgo\_web\_search} consistently fails due to rate limiting (0/38 success; repeated ``anomaly'' warnings), and these failures are persistently recorded by the quality manager.
As a result, later retrieval progressively deprioritizes this tool until it drops out of the top-$K$ candidate set.

This shift is visible in two news-seeking tasks:
\textbf{In Task 20} ("today's top LLM-related news"), \texttt{duckduckgo\_web\_search} is highly ranked by semantic similarity and is invoked once, immediately failing and wasting $\sim$3s latency; the task completes only because alternative news tools (\textit{e.g.}, \texttt{get-nytimes-news}, \texttt{get-theverge-news}, \texttt{get-infoq-news}) provide sufficient coverage.
\textbf{In Task 77} ("big AI news today"), \texttt{duckduckgo\_web\_search} is demoted outside top-$K$ (rank $1\!\rightarrow\!18$, $K{=}15$) and thus never selected; the agent instead uses reliable sources (\textit{e.g.}, \texttt{get-infoq-news}, \texttt{get-theverge-news}, \texttt{get-bbc-news}, \texttt{get-36kr-trending}) to complete tasks with zero failed calls. Overall, self-evolution yields \emph{graceful degradation}: the agent automatically stops attempting broken tools while maintaining task success and reducing wasted executions.

\WFclear

\Needspace{18\baselineskip}
\begin{wrapfigure}[15]{r}{0.50\textwidth}
    \vspace{\dimexpr-\intextsep+10pt\relax}
    \centering
    \includegraphics[width=\linewidth,trim=0 10bp 0 0,clip]{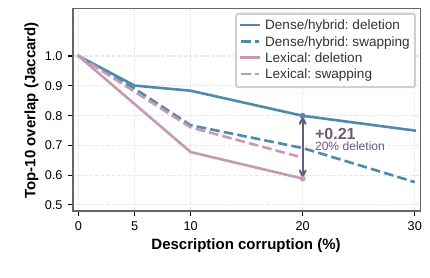}
    \captionsetup{font=normalsize,skip=10pt}
    \caption{Top-10 retrieval stability under description noise. \textcolor{blue}{Blue}: dense/hybrid; \textcolor{green!60!black}{purple}: lexical. Solid: deletion; dashed: swapping.}
    \label{fig:metadata_stability}

\end{wrapfigure}

\textbf{Robustness to description noise.} Tool metadata is often incomplete or misleading in practice. We perturb the descriptions of the 109 cached OSMCP tools by deleting 5--30\% of words or swapping descriptions across tools, keeping names and servers intact, and retrieve for 18 application queries (Figure~\ref{fig:metadata_stability}; Appendix~\ref{app:metadata_protocol}). At $K=10$, the dense/hybrid ranker retains 0.75 Top-10 overlap with clean retrieval under 30\% deletion and 0.58 under 30\% swapping, while its expected-server coverage stays at 18/18 throughout; the lexical control degrades faster (0.59 at 20\% deletion). The right application's tools thus remain in the action space even when fine-grained descriptions are corrupted.

\WFclear

% \subsection{Efficiency Comparison}

\setcounter{topnumber}{2}

\vspace{-0.1in}
\section{Related Work}
\vspace{-0.05in}
\textbf{Tool-Augmented LLM Agents.} The growing capability of large language models (LLMs) to use tools has attracted significant research interest, as it greatly extends their practical utility~\citep{shi2025retrieval}. Early approaches~\citep{press2023measuring,yao2022react} relied on prompting strategies that interleaved reasoning with external queries or API calls, while others~\citep{schick2023toolformer} proposed training-based methods to equip LLMs with tool-using skills. Following the emergence of open protocols like the Model Context Protocol (MCP)~\citep{mcp}, the tool ecosystem has shifted from closed to open systems. This shift introduces new challenges, including an explosive growth in the number of available tools and significant variation in their quality---issues that traditional context-only methods handle poorly~\citep{shen2023hugginggpt}. Recent efforts aim to address these problems. For example, AutoTool~\citep{zou2025autotool} improves model adaptability to evolving tool sets through multi-stage post-training, while ToolACE-MCP~\citep{yao2026toolace} enables precise routing within large-scale tool ecosystems by constructing trajectory graphs.

\textbf{Computer-Using Agents.} Advances in vision-language models and expanding agent applications have driven the development of computer-using agents, which are designed to perform diverse and complex tasks on computers~\citep{anthropic, xie2024osworld}. The diversity and complexity of these tasks, along with their need for robust GUI-grounded capabilities, expose limitations in traditional tool-based approaches. Current research typically depends either on a model's inherent GUI proficiency~\citep{lu2024omniparser, claude-4, wang2025ui} or on agentic frameworks~\citep{yang2026symphony, song2025coact, xie2025scaling}. For instance, UI-TARS~\citep{qin2025ui} achieves strong GUI task performance via multi-stage post-training on targeted, high-quality data, whereas Agent S3~\citep{gonzalez2025unreasonable} employs a framework that uses code-based actions to improve task success rates. With the growth of the MCP ecosystem, a promising direction is to integrate MCP's efficient interfacing with GUI interaction and more fine-grained, flexible, and generalizable tools to collaboratively solve computer-using tasks.

\vspace{-0.1in}
\section{Conclusion}
\vspace{-0.05in}

We introduce \model, a universal action layer for self-evolving agents operating in open-world tool ecosystems, designed to handle scale, non-stationarity, and heterogeneity. By combining hierarchical progressive retrieval, test-time reliability evolution, and cross-modal observation grounding, \model adaptively builds and evolves its action space from execution feedback and robustly orchestrates actions, balancing both task success and efficiency. On LiveMCPBench and OSMCP, it achieves SOTA effectiveness and efficiency with modest base-LLM requirements.

\clearpage

\bibliographystyle{iclr2027_conference}
\bibliography{refs}

\clearpage

\appendix
\begingroup
\newsavebox{\AnyActAppendixTableBox}
\newcommand{\AnyActSummaryTableStyle}{%
    \fontsize{8}{8.5}\selectfont
    \renewcommand{\arraystretch}{1}%
    \setlength{\tabcolsep}{3pt}%
    \setlength{\aboverulesep}{1.5pt}%
    \setlength{\belowrulesep}{1.5pt}%
}
\raggedbottom
\captionsetup[table]{font=small,skip=4pt,justification=raggedright,singlelinecheck=true}
\setlength{\textfloatsep}{8pt plus 1pt minus 1pt}
\setlength{\floatsep}{6pt plus 1pt minus 1pt}
\setlength{\intextsep}{6pt plus 1pt minus 1pt}
\renewcommand{\arraystretch}{1.06}
\setlength{\LTpre}{8pt}
\setlength{\LTpost}{8pt}
\setlength{\LTcapwidth}{\textwidth}
\section{OSMCP Benchmark Details}
\label{app:osmcp}

\subsection{Task Selection and Outcome Evaluation}
\label{app:osmcp_tasks}
OSMCP contains all 198 tasks in the seven application domains listed
in Table~\ref{tab:osmcp_coverage}, with no additional filtering within
those domains. This selection follows application-server coverage and
represents a subset of OSWorld, excluding its dedicated Chrome, OS,
and multi-application task groups. Comparison against the corresponding
base snapshot confirms that all 198 task instructions and evaluator
specifications are unchanged. The 17 VLC tasks differ only in runtime
configuration and explanatory notes. Their initial-state requirements
use either Chrome or base-setup VM snapshots, illustrating that task
domains and runtime images serve different roles in the benchmark.

\begin{table}[htbp]
\centering
\caption{OSMCP task coverage and static tool inventory. Infeasible tasks
are included in the task counts.}
\label{tab:osmcp_coverage}
\AnyActSummaryTableStyle
\begin{tabular}{l|cc|cccc}
\toprule
\multicolumn{1}{c|}{\textbf{Application}} & \textbf{Tasks} & \textbf{Infeasible} & \textbf{Tools} & \textbf{Atomic} & \textbf{Compound} & \textbf{Workflow} \\
\midrule
LibreOffice Calc    & 47 & 1  & 32 & 6  & 21 & 5 \\
LibreOffice Impress & 47 & 0  & 23 & 3  & 20 & 0 \\
LibreOffice Writer  & 23 & 1  & 21 & 5  & 14 & 2 \\
GIMP                & 26 & 10 & 9  & 4  & 4  & 1 \\
Thunderbird         & 15 & 1  & 13 & 10 & 3  & 0 \\
VLC                 & 17 & 3  & 11 & 8  & 3  & 0 \\
VS Code             & 23 & 5  & 13 & 9  & 4  & 0 \\
\midrule
\rowcolor{gray!25}
\textbf{Total} & \textbf{198} & \textbf{21} & \textbf{122} & \textbf{45} & \textbf{69} & \textbf{8} \\
\bottomrule
\end{tabular}
\end{table}

The inherited executable evaluators assess application state and
produced artifacts, including spreadsheets, documents, presentations,
images, and configuration files. Task scores thus measure the final
outcome beyond the success status of individual tool calls.
The task set also includes 21 cases with the \texttt{infeasible}
evaluator. These receive a score of one when the final action is
\texttt{FAIL}, including its structured-action form, and zero otherwise.
They remain part of the 198-task denominator, allowing the evaluation
to capture recognition of infeasibility. Table~\ref{tab:exp_OSWorld}
includes the available task count for each system: Agentic-Lybic-Maestro
and GBOX Agent have 197 outcomes, while the other rows cover 198 tasks.

\subsection{Tool Interfaces and Granularity}
\label{app:osmcp_granularity}
The static tool catalog is annotated along two dimensions.
Operation type describes functional intent through seven categories:
Read, Write, Create, Transform, Control, Export, and Configuration.
Granularity describes the scope of the exposed interface.
Atomic tools provide individual application-level operations, such as
\texttt{get\_zoom} and \texttt{create\_new\_sheet}.
Compound tools compose operations or process data in bulk;
\texttt{Fill\_blank\_above}, for instance, fills blank cells using
values from preceding cells, while \texttt{calculate\_employee\_ages}
transforms a birthday column into ages.
Workflow tools implement broader procedures, exemplified by
\texttt{calculate\_annual\_asset\_changes} and
\texttt{summarize\_revenue\_by\_promotion}.
Table~\ref{tab:CoWork-MCP stat} summarizes both dimensions by application.

\begin{table}[htbp]
\centering
\caption{Operation types and interface granularity in the OSMCP tool catalog.}
\label{tab:CoWork-MCP stat}
\AnyActSummaryTableStyle
\setlength{\tabcolsep}{2pt}
\begin{tabular}{l|ccccccc|c}
\toprule
\multicolumn{1}{c|}{\textbf{Type}} & \textbf{Calc} & \textbf{Impress} & \textbf{Writer} & \textbf{Thunderbird} & \textbf{VS Code} & \textbf{VLC} & \textbf{GIMP} & \textbf{Total} \\
\midrule
\rowcolor{gray!10}
\multicolumn{9}{c}{\textit{Operation type}} \\
Read      & 3 & 3 & 1 & 4 & 5 & 3 & 1 & 20 \\
Write     & 8 & 13 & 15 & 3 & 1 & 1 & 1 & 42 \\
Create    & 6 & 2 & 1 & 3 & 0 & 0 & 0 & 12 \\
Transform & 9 & 0 & 3 & 0 & 0 & 0 & 5 & 17 \\
Control   & 3 & 1 & 0 & 0 & 7 & 6 & 1 & 18 \\
Export    & 2 & 3 & 1 & 0 & 0 & 1 & 1 & 8 \\
Configuration & 1 & 1 & 0 & 3 & 0 & 0 & 0 & 5 \\
\midrule
\rowcolor{gray!10}
\multicolumn{9}{c}{\textit{Granularity}} \\
Atomic   & 6 & 3 & 5 & 10 & 9 & 8 & 4 & 45 \\
Compound & 21 & 20 & 14 & 3 & 4 & 3 & 4 & 69 \\
Workflow & 5 & 0 & 2 & 0 & 0 & 0 & 1 & 8 \\
\midrule
\rowcolor{gray!25}
\textbf{Total} & \textbf{32} & \textbf{23} & \textbf{21} & \textbf{13} & \textbf{13} & \textbf{11} & \textbf{9} & \textbf{122} \\
\bottomrule
\end{tabular}
\end{table}

These annotations characterize the functional scope presented to the
agent. Internal execution may involve several low-level operations
even for an Atomic interface, and a general dispatch interface may
support a wide range of behaviors. Consequently, execution cost and
task difficulty remain separate properties of a tool invocation.

\subsection{Benchmark Construction and Generalization}
\label{app:osmcp_scope}
The catalog was developed in the context of publicly available
OSWorld tasks. Its reusable application controls support diverse
goals, while specialized procedures expose useful operations for
particular subtasks. For example, \texttt{calculate\_employee\_ages}
is a Compound interface tailored to an age-calculation operation.
Task specificity thus cuts across the three granularity levels:
it depends on an interface's semantics, beyond the number or breadth
of operations it exposes. The eight Workflow annotations capture
only one part of this specialization.

OSMCP measures how agents use this mixture of interfaces on the
covered tasks. Performance reflects both the available tool support
and the agent's selection and coordination of tools. Because the
tasks were public during catalog development, tool design can
incorporate knowledge of the evaluation setting; prior model exposure
and prompt or parameter tuning are further possible sources of
adaptation, including for a training-free controller. The present
results characterize this setting. Transfer to held-out task families
or applications with a frozen catalog, and the effect of removing
task-aligned interfaces across granularity levels, remain open
empirical questions.

\subsection{Benchmark Artifacts and Runtime Dependencies}
\label{app:osmcp_reproducibility}
The accompanying task manifest
(\path{benchmark/osmcp_task_ids.json}) identifies the selected tasks,
and the annotated catalog
(\path{benchmark/osmcp_tool_catalog.csv}) records the static interfaces.
The definition metadata
(\path{benchmark/osmcp_definition.json}) supplies the base OSWorld
snapshot identifier and file hashes. These artifacts specify the
static task and tool inventory. Its realization in a running system
depends on server availability, tool schemas, application versions,
dependencies, VM snapshots, input assets, and initialization settings.
The interfaces available to an individual run can therefore differ
from the static catalog. The definition files record benchmark
coverage rather than every historical execution environment.

Execution histories are maintained per backend, server, and action name
and retained across tasks. The implementation defaults to a window of
100 calls, a three-call cold start, a success-rate threshold of 0.4, and a
minimum reliability multiplier of 0.2. Independently of the success-rate term,
the consecutive-failure deduction is 0.1 at the third failure and increases
by 0.1 per additional failure, capped at 0.3. These histories track
the runtime execution-status signal defined in Section~\ref{sec:method}.

Table~\ref{tab:exp_OSWorld} presents the available systems with their
native models, interfaces, and step budgets. A step may denote an
agent decision, an MCP invocation, or a GUI action, with additional
operations executed inside tools or shell commands. The resulting
scores describe performance at the system level. Isolating the
controller's contribution calls for a shared model and catalog under
matched execution policies, while time and token measurements provide
separate measures of cost.

\subsection{Description-Noise Evaluation Setup}
\label{app:metadata_protocol}
Figure~\ref{fig:metadata_stability} uses 18 queries, three per application,
over 109 cached tools from six OSMCP applications. With names and server
identities fixed, descriptions undergo deterministic word deletion or cyclic
swapping at nominal rates of 5/10/20/30\%; the lexical control covers 10/20\%.
We average the Jaccard overlap between clean and perturbed Top-10 tool-name
sets over the queries, using one perturbation realization per rate.
Expected-server coverage is the fraction of queries whose Top-10 contains
at least one tool from the expected application. Dense and hybrid rankings
coincide because the hybrid prefilter's 120-candidate limit retains all
109 tools; the lexical control uses normalized query-term overlap. This
evaluation tests local ranking, without server planning or task execution.

\section{Complete List of MCP Tools}
\label{app:osmcp_tools}
% \section{Complete List of MCP Tools}\label{apd:mcp_tools}

We provide the complete list of 122 MCP tools across seven applications in Tables~\ref{tab:mcp_calc}--\ref{tab:mcp_gimp}.

\begingroup
\scriptsize
\setlength{\LTpre}{10pt}
\setlength{\LTpost}{10pt}
\setlength{\tabcolsep}{2.6pt}
\renewcommand{\arraystretch}{1.18}
\begin{longtable}{@{}>{\raggedright\arraybackslash}p{0.35\textwidth}>{\raggedright\arraybackslash}p{0.38\textwidth}>{\raggedright\arraybackslash}p{0.13\textwidth}>{\raggedright\arraybackslash}p{0.10\textwidth}@{}}
\caption{MCP tools for \textbf{Calc} (32 tools).}\label{tab:mcp_calc} \\[6pt]
\toprule
Tool Name & Description & Operation Type & Granularity \\
\midrule
\endfirsthead
\multicolumn{4}{l}{\textit{Table \thetable\ (continued)}} \\[4pt]
\toprule
Tool Name & Description & Operation Type & Granularity \\
\midrule
\endhead
\bottomrule
\endfoot
\path{Fill_blank_above} & Fill blank cells with values from cells above & Write & Compound \\
\path{add_new_row} & Add new row with data and formulas & Write & Compound \\
\path{calculate_annual_asset_changes} & Calculate year-over-year changes in asset values & Transform & Workflow \\
\path{calculate_employee_ages} & Calculate ages from birthday column & Transform & Compound \\
\path{clean_text_formatting} & Clean and format text in columns (remove whitespace, apply title case) & Transform & Compound \\
\path{copy_column} & Copy a column to a new sheet & Write & Compound \\
\path{create_dropdown_validation} & Create dropdown validation for column & Create & Compound \\
\path{create_line_chart} & Create line chart from data & Create & Workflow \\
\path{create_new_sheet} & Create a new empty sheet & Create & Atomic \\
\path{create_pivot_table} & Create pivot table from column data & Create & Workflow \\
\path{create_sheet_with_headers} & Create new sheet with merged header cells & Create & Compound \\
\path{export_csv} & Export sheet to CSV format & Export & Compound \\
\path{export_pdf} & Export spreadsheet to PDF with custom sizing & Export & Compound \\
\path{extract_unique_values} & Extract unique values from a column to another column & Transform & Compound \\
\path{fill_sequence_numbers} & Fill column with formatted sequence numbers & Write & Compound \\
\path{format_column_decimals} & Format column with decimal places and currency symbols & Write & Compound \\
\path{freeze_panes} & Freeze rows and/or columns & Control & Atomic \\
\path{get_all_document_contents} & Read all data from document & Read & Compound \\
\path{get_zoom} & Get current spreadsheet zoom level & Read & Atomic \\
\path{hide_rows_with_value} & Hide or show rows containing specific values & Control & Compound \\
\path{highlight_weekends} & Highlight weekend dates with background color & Write & Compound \\
\path{manage_sheets} & Perform sheet operations (rename, copy) & Write & Compound \\
\path{mcp_libreoffice_status} & Check LibreOffice setup status and get installation instructions & Read & Atomic \\
\path{pad_numbers} & Format numbers with custom padding (e.g., 123 $\rightarrow$ 0000123) & Transform & Compound \\
\path{reorder_columns} & Reorder columns in specified sequence & Transform & Compound \\
\path{set_cell_content} & Set content in specific cells with formatting & Write & Atomic \\
\path{set_locale_formatting} & Set locale-specific number formatting & Configuration & Compound \\
\path{set_zoom} & Set spreadsheet zoom level & Control & Atomic \\
\path{setup_resource_sheets} & Set up resource management sheets with rename, backup, and offline version & Create & Workflow \\
\path{sort_data} & Sort spreadsheet data by column & Transform & Compound \\
\path{summarize_revenue_by_promotion} & Create revenue summary grouped by promotion type & Transform & Workflow \\
\path{transpose_table} & Transpose table data (rows $\leftrightarrow$ columns) & Transform & Compound \\
\end{longtable}
\endgroup

\begingroup
\scriptsize
\setlength{\LTpre}{10pt}
\setlength{\LTpost}{10pt}
\setlength{\tabcolsep}{2.6pt}
\renewcommand{\arraystretch}{1.18}
\begin{longtable}{@{}>{\raggedright\arraybackslash}p{0.35\textwidth}>{\raggedright\arraybackslash}p{0.38\textwidth}>{\raggedright\arraybackslash}p{0.13\textwidth}>{\raggedright\arraybackslash}p{0.10\textwidth}@{}}
\caption{MCP tools for \textbf{Impress} (23 tools).}\label{tab:mcp_impress} \\[6pt]
\toprule
Tool Name & Description & Operation Type & Granularity \\
\midrule
\endfirsthead
\multicolumn{4}{l}{\textit{Table \thetable\ (continued)}} \\[4pt]
\toprule
Tool Name & Description & Operation Type & Granularity \\
\midrule
\endhead
\bottomrule
\endfoot
\path{mcp_libreoffice_add_image_to_slides} & Add images to slides with configurable size and position & Write & Compound \\
\path{mcp_libreoffice_add_slide_notes} & Add notes to slides, either custom text or copying the slide's title text & Write & Compound \\
\path{mcp_libreoffice_analyze_slides} & Analyze and extract all text contents from slides with formatting info & Read & Compound \\
\path{mcp_libreoffice_configure_auto_save} & Configure auto-save functionality to automatically save every N minutes & Configuration & Atomic \\
\path{mcp_libreoffice_create_new_slides} & Create new blank slides with configurable positioning & Create & Compound \\
\path{mcp_libreoffice_duplicate_slides} & Duplicate specific slides or the last N slides & Create & Compound \\
\path{mcp_libreoffice_export_to_pdf} & Export presentation to PDF with full quality control & Export & Compound \\
\path{mcp_libreoffice_export_to_png} & Export presentation to PNG with flexible resolution settings & Export & Compound \\
\path{mcp_libreoffice_get_slide_contents} & Extract and analyze all text contents from slides & Read & Compound \\
\path{mcp_libreoffice_reposition_images} & Reposition existing images with predefined or custom positions & Write & Compound \\
\path{mcp_libreoffice_reposition_title} & Reposition title text on slides (top, middle, bottom) & Write & Compound \\
\path{mcp_libreoffice_restore_slide_panel} & Restore the left slide panel when accidentally closed & Control & Atomic \\
\path{mcp_libreoffice_save_as_powerpoint} & Save presentation as PowerPoint (.pptx/.ppt) & Export & Compound \\
\path{mcp_libreoffice_set_font_styles} & Set font style (font family) for text in specific slides & Write & Compound \\
\path{mcp_libreoffice_set_slide_number_color} & Set color for slide numbers & Write & Compound \\
\path{mcp_libreoffice_set_slide_orientation} & Set slide orientation (portrait/landscape) & Write & Compound \\
\path{mcp_libreoffice_set_text_bold} & Set text boldness with selective targeting (all/titles/content) & Write & Compound \\
\path{mcp_libreoffice_set_text_colors} & Set text colors for all textboxes in slides & Write & Compound \\
\path{mcp_libreoffice_set_title_colors} & Set title text color for specific slides & Write & Compound \\
\path{mcp_libreoffice_set_title_font_size} & Set title font size for specific slides & Write & Compound \\
\path{mcp_libreoffice_set_title_text} & Set title text for specific slides & Write & Compound \\
\path{mcp_libreoffice_set_title_underlines} & Set title text underline for specific slides & Write & Compound \\
\path{mcp_libreoffice_status} & Check LibreOffice setup status and get installation instructions & Read & Atomic \\
\end{longtable}
\endgroup

\begingroup
\scriptsize
\setlength{\LTpre}{10pt}
\setlength{\LTpost}{10pt}
\setlength{\tabcolsep}{2.6pt}
\renewcommand{\arraystretch}{1.18}
\begin{longtable}{@{}>{\raggedright\arraybackslash}p{0.35\textwidth}>{\raggedright\arraybackslash}p{0.38\textwidth}>{\raggedright\arraybackslash}p{0.13\textwidth}>{\raggedright\arraybackslash}p{0.10\textwidth}@{}}
\caption{MCP tools for \textbf{Writer} (21 tools).}\label{tab:mcp_writer} \\[6pt]
\toprule
Tool Name & Description & Operation Type & Granularity \\
\midrule
\endfirsthead
\multicolumn{4}{l}{\textit{Table \thetable\ (continued)}} \\[4pt]
\toprule
Tool Name & Description & Operation Type & Granularity \\
\midrule
\endhead
\bottomrule
\endfoot
\path{mcp_libreoffice_add_content_at_cursor} & Add content at the current cursor position & Write & Atomic \\
\path{mcp_libreoffice_add_page_numbers} & Add page numbers to the document & Write & Compound \\
\path{mcp_libreoffice_append_content} & Append plain text content to the document & Write & Atomic \\
\path{mcp_libreoffice_append_formatted_content} & Append formatted content to the document & Write & Compound \\
\path{mcp_libreoffice_change_font} & Change font throughout the entire document text & Write & Compound \\
\path{mcp_libreoffice_conditional_text_formatting} & Apply formatting based on text conditions (e.g., color words by pattern) & Write & Workflow \\
\path{mcp_libreoffice_data_processing} & Process structured data (remove duplicates, filter, sort, etc.) & Transform & Workflow \\
\path{mcp_libreoffice_export_pdf} & Export document to PDF & Export & Compound \\
\path{mcp_libreoffice_insert_image} & Insert an image at the current cursor position & Write & Compound \\
\path{mcp_libreoffice_insert_table} & Insert a table at the current cursor position & Create & Compound \\
\path{mcp_libreoffice_italic_text_formatting} & Modify italic text formatting for better discernibility & Write & Compound \\
\path{mcp_libreoffice_line_spacing} & Modify line spacing for specified text portions & Write & Compound \\
\path{mcp_libreoffice_page_break} & Insert page breaks to create blank pages or separate content & Write & Atomic \\
\path{mcp_libreoffice_paragraph_formatting} & Apply formatting to entire paragraphs (strikethrough, bold, etc.) & Write & Compound \\
\path{mcp_libreoffice_remove_highlighting} & Remove all highlighting/background colors from text & Write & Compound \\
\path{mcp_libreoffice_replace_formatted_text} & Replace text with formatted content & Write & Compound \\
\path{mcp_libreoffice_replace_text} & Replace text in the document & Write & Atomic \\
\path{mcp_libreoffice_sentence_separation} & Separate sentences by adding empty lines between them & Transform & Compound \\
\path{mcp_libreoffice_status} & Check LibreOffice setup status and get installation instructions & Read & Atomic \\
\path{mcp_libreoffice_text_alignment} & Change text alignment (left, center, right, justify) & Write & Compound \\
\path{mcp_libreoffice_text_case_transformation} & Transform text case (uppercase, lowercase, title case, etc.) & Transform & Compound \\
\end{longtable}
\endgroup

\begingroup
\scriptsize
\setlength{\LTpre}{10pt}
\setlength{\LTpost}{10pt}
\setlength{\tabcolsep}{2.6pt}
\renewcommand{\arraystretch}{1.18}
\begin{longtable}{@{}>{\raggedright\arraybackslash}p{0.35\textwidth}>{\raggedright\arraybackslash}p{0.38\textwidth}>{\raggedright\arraybackslash}p{0.13\textwidth}>{\raggedright\arraybackslash}p{0.10\textwidth}@{}}
\caption{MCP tools for \textbf{Thunderbird} (13 tools).}\label{tab:mcp_thunderbird} \\[6pt]
\toprule
Tool Name & Description & Operation Type & Granularity \\
\midrule
\endfirsthead
\multicolumn{4}{l}{\textit{Table \thetable\ (continued)}} \\[4pt]
\toprule
Tool Name & Description & Operation Type & Granularity \\
\midrule
\endhead
\bottomrule
\endfoot
\path{add_email_attachment} & Add an attachment to a compose window & Write & Atomic \\
\path{bulk_flag_folder} & Flag all messages in a specific folder & Write & Compound \\
\path{get_accounts} & Get email accounts via API & Read & Atomic \\
\path{get_auto_forward_rules} & Get all configured automatic email forwarding rules & Read & Atomic \\
\path{get_folders} & Get folders via API & Read & Atomic \\
\path{get_subject_filter_rules} & Get all configured subject-based email filtering rules & Read & Atomic \\
\path{mcp_thunderbird_create_folder} & Create a new folder via API & Create & Atomic \\
\path{mcp_thunderbird_new_identity} & Create a new mail identity via API & Create & Atomic \\
\path{mcp_thunderbird_theme_change} & Change the theme/appearance & Configuration & Atomic \\
\path{open_compose_window} & Open a new email compose window & Create & Atomic \\
\path{setup_auto_forward} & Set up automatic email forwarding rule & Configuration & Compound \\
\path{setup_subject_filter} & Set up automatic email filtering based on subject keywords & Configuration & Compound \\
\path{update_message} & Update a message via API & Write & Atomic \\
\end{longtable}
\endgroup

\begingroup
\scriptsize
\setlength{\LTpre}{10pt}
\setlength{\LTpost}{10pt}
\setlength{\tabcolsep}{2.6pt}
\renewcommand{\arraystretch}{1.18}
\begin{longtable}{@{}>{\raggedright\arraybackslash}p{0.35\textwidth}>{\raggedright\arraybackslash}p{0.38\textwidth}>{\raggedright\arraybackslash}p{0.13\textwidth}>{\raggedright\arraybackslash}p{0.10\textwidth}@{}}
\caption{MCP tools for \textbf{VS Code} (13 tools).}\label{tab:mcp_vscode} \\[6pt]
\toprule
Tool Name & Description & Operation Type & Granularity \\
\midrule
\endfirsthead
\multicolumn{4}{l}{\textit{Table \thetable\ (continued)}} \\[4pt]
\toprule
Tool Name & Description & Operation Type & Granularity \\
\midrule
\endhead
\bottomrule
\endfoot
\path{code_checker} & Retrieve diagnostics from language services for the active workspace & Read & Atomic \\
\path{execute_command} & Execute a command in an integrated terminal & Control & Compound \\
\path{execute_vscode_command} & Execute any VSCode command by its command ID & Control & Atomic \\
\path{focus_editor} & Open file in editor and navigate to specific line and column & Control & Atomic \\
\path{get_terminal_output} & Retrieve the output from a specific terminal by ID & Read & Atomic \\
\path{list_debug_sessions} & List all active debug sessions in the workspace & Read & Atomic \\
\path{list_directory} & List directory contents in a tree format & Read & Atomic \\
\path{list_vscode_commands} & List available VSCode commands with optional filtering & Read & Atomic \\
\path{preview_url} & Open a URL in VSCode's built-in simple browser & Control & Atomic \\
\path{restart_debug_session} & Restart a debug session with the provided configuration & Control & Compound \\
\path{start_debug_session} & Start a new debug session with the provided configuration & Control & Compound \\
\path{stop_debug_session} & Stop all debug sessions matching the provided session name & Control & Atomic \\
\path{text_editor} & File manipulation using VSCode's native APIs & Write & Compound \\
\end{longtable}
\endgroup

\begingroup
\scriptsize
\setlength{\LTpre}{10pt}
\setlength{\LTpost}{10pt}
\setlength{\tabcolsep}{2.6pt}
\renewcommand{\arraystretch}{1.18}
\begin{longtable}{@{}>{\raggedright\arraybackslash}p{0.35\textwidth}>{\raggedright\arraybackslash}p{0.38\textwidth}>{\raggedright\arraybackslash}p{0.13\textwidth}>{\raggedright\arraybackslash}p{0.10\textwidth}@{}}
\caption{MCP tools for \textbf{VLC} (11 tools).}\label{tab:mcp_vlc} \\[6pt]
\toprule
Tool Name & Description & Operation Type & Granularity \\
\midrule
\endfirsthead
\multicolumn{4}{l}{\textit{Table \thetable\ (continued)}} \\[4pt]
\toprule
Tool Name & Description & Operation Type & Granularity \\
\midrule
\endhead
\bottomrule
\endfoot
\path{add_url_to_playlist} & Add a URL to playlist without playing immediately & Write & Atomic \\
\path{adjust_volume} & Increase or decrease the volume by a specified percentage & Control & Atomic \\
\path{get_available_videos} & Get all available videos with their path & Read & Atomic \\
\path{get_status} & Get the current status of playback & Read & Atomic \\
\path{get_volume} & Get the current volume level (as a percentage) & Read & Atomic \\
\path{seek} & Seek to a specific position in the video & Control & Atomic \\
\path{set_volume} & Set the volume level (0--200, where 100 is normal) & Control & Atomic \\
\path{show_video} & Show the video using path and subtitle language code & Control & Compound \\
\path{stream_url} & Stream a video/audio URL using the HTTP API & Control & Compound \\
\path{take_video_snapshot} & Take a snapshot of the current video scene & Export & Compound \\
\path{vlc_control} & Control playback with actions: play, pause, stop, fullscreen & Control & Atomic \\
\end{longtable}
\endgroup

\begingroup
\scriptsize
\setlength{\LTpre}{10pt}
\setlength{\LTpost}{10pt}
\setlength{\tabcolsep}{2.6pt}
\renewcommand{\arraystretch}{1.18}
\begin{longtable}{@{}>{\raggedright\arraybackslash}p{0.35\textwidth}>{\raggedright\arraybackslash}p{0.38\textwidth}>{\raggedright\arraybackslash}p{0.13\textwidth}>{\raggedright\arraybackslash}p{0.10\textwidth}@{}}
\caption{MCP tools for \textbf{GIMP} (9 tools).}\label{tab:mcp_gimp} \\[6pt]
\toprule
Tool Name & Description & Operation Type & Granularity \\
\midrule
\endfirsthead
\multicolumn{4}{l}{\textit{Table \thetable\ (continued)}} \\[4pt]
\toprule
Tool Name & Description & Operation Type & Granularity \\
\midrule
\endhead
\bottomrule
\endfoot
\path{adjust_brightness_contrast} & Adjust the brightness and contrast of an image or layer & Transform & Atomic \\
\path{call_api} & Call GIMP API through the socket connection & Control & Workflow \\
\path{convert_to_palette_mode} & Convert an image to palette-based (indexed color) mode & Transform & Compound \\
\path{enhance_color_vibrancy} & Enhance color vibrancy and saturation using various methods & Transform & Compound \\
\path{export_image} & Export an image to a specified location with a given filename & Export & Compound \\
\path{flip_image} & Flip an image or layer horizontally or vertically & Transform & Atomic \\
\path{get_images} & Get information about currently open images & Read & Atomic \\
\path{make_background_transparent} & Make the background transparent using various methods & Transform & Compound \\
\path{move_text_layer} & Move a text layer in the specified direction & Write & Atomic \\
\end{longtable}
\endgroup

\endgroup

\end{document}